\documentclass[11pt,a4paper]{article}
\usepackage[hyperref]{acl2020}
\usepackage{times}
\usepackage{latexsym}

\usepackage{microtype}
\usepackage{graphicx}
\usepackage{subfig}
\usepackage{booktabs} 
\usepackage{pifont}
\usepackage{hyperref}

\usepackage{kevinlin}
\usepackage{mingyul}

\usepackage{times}
\usepackage{latexsym}
\usepackage[export]{adjustbox}
\usepackage{multirow}
\usepackage{booktabs}
\usepackage{graphicx}
\usepackage{epstopdf}
\usepackage{epsfig}
\usepackage{amssymb}
\usepackage{amsmath}
\usepackage{comment}
\usepackage{tabularx}
\usepackage{enumitem}
\usepackage{comment}
\usepackage{color}
\usepackage{sidecap}
\usepackage{url}

\usepackage{pifont}

\usepackage{url}

\aclfinalcopy 
\def\aclpaperid{959} 

\title{Learning to Generate Multiple Style Transfer Outputs \\
for an Input Sentence}

\author{Kevin Lin\textsuperscript{\ding{168}}, Ming-Yu Liu\textsuperscript{\ding{169}}, Ming-Ting Sun\textsuperscript{\ding{168}}, Jan Kautz\textsuperscript{\ding{169}} \\
            \textsuperscript{\ding{168}}University of Washington
            \textsuperscript{\ding{169}}NVIDIA \\
            {\small \tt \{kvlin,mts\}@uw.edu, \{mingyul,jkautz\}@nvidia.com}
}
\date{}

\begin{document}
\maketitle

\begin{abstract}
Text style transfer refers to the task of rephrasing a given text in a different style. While various methods have been proposed to advance the state of the art, they often assume the transfer output follows a delta distribution, and thus their models cannot generate different style transfer results for a given input text. To address the limitation, we propose a one-to-many text style transfer framework. In contrast to prior works that learn a one-to-one mapping that converts an input sentence to one output sentence, our approach learns a one-to-many mapping that can convert an input sentence to multiple different output sentences, while preserving the input content. This is achieved by applying adversarial training with a latent decomposition scheme. Specifically, we decompose the latent representation of the input sentence to a style code that captures the language style variation and a content code that encodes the language style-independent content. We then combine the content code with the style code for generating a style transfer output. By combining the same content code with a different style code, we generate a different style transfer output. Extensive experimental results with comparisons to several text style transfer approaches on multiple public datasets using a diverse set of performance metrics validate effectiveness of the proposed approach.
\end{abstract}

\section{Introduction}

Text style transfer aims at changing the language style of an input sentence to a target style with the constraint that the style-independent content should remain the same across the transfer. While several methods are proposed for the task~\cite{john2019disentangled,smith2019zero,jhamtani2017shakespearizing,ker1992style,xu2012paraphrasing,shen2017style,subramanian2018multiple,xu2018unpaired}, they commonly model the distribution of the transfer outputs as a delta distribution, which implies a one-to-one mapping mechanism that converts an input sentence in one language style to a \emph{single} corresponding sentence in the target language style. 

We argue a multimodal mapping is better suited for the text style transfer task. For examples, the following two reviews: 
\begin{enumerate}[noitemsep,nolistsep,leftmargin=*]
	\item ``\it{This lightweight vacuum is simply effective.}'', 
	\item ``\it{This easy-to-carry vacuum picks up dust and trash amazingly well.}''
\end{enumerate}
would both be considered correct negative-to-positive transfer results for the input sentence, ``\textit{This heavy vacuum sucks}''. Furthermore, a one-to-many mapping allows a user to pick the preferred text style transfer outputs in the inference time.

\begin{figure*}[t]
	\centering
	\includegraphics[width=.85\textwidth]{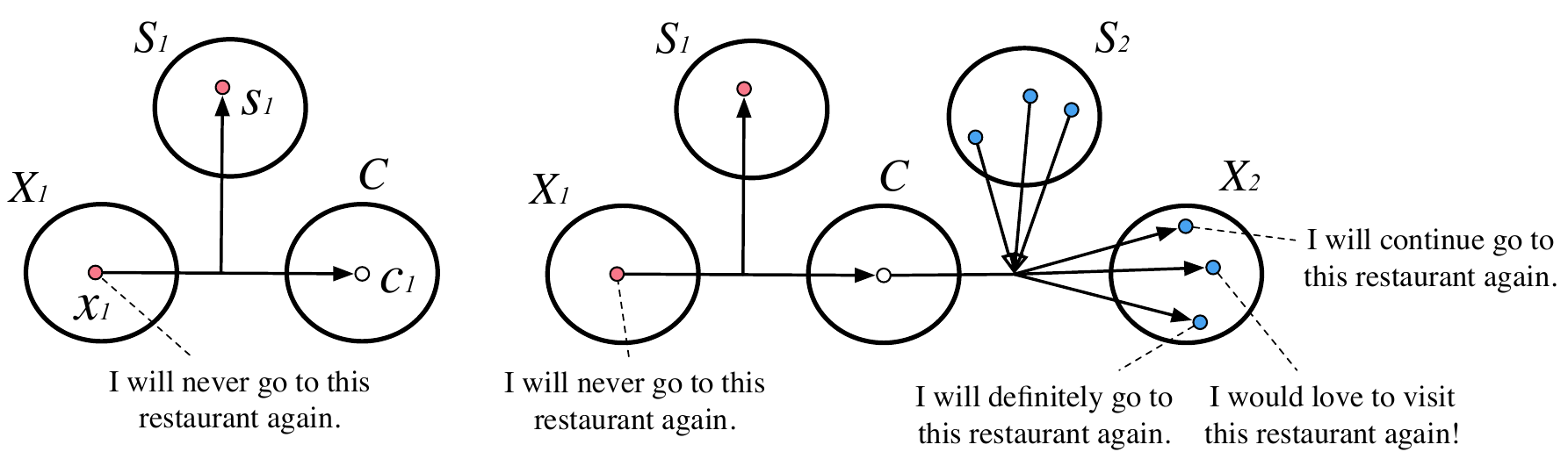}
	\caption{We formulate text style transfer as a one-to-many mapping function. Left: We decompose the sentence $x_1$ to a content code $c_1$ that controls the sentence meaning, and a style code $s_1$ that captures the stylistic properties of the input $x_1$. Right: One-to-many style transfer is achieved by fusing the content code $c_1$ and a style code $s_2$ randomly sampled from the target style space $S_2$.}
	\label{fig:problem}
	\vspace{-3mm}
\end{figure*}

In this paper, we propose a one-to-many text style transfer framework that can be trained using non-parallel text. That is, we assume the training data consists of two corpora of different styles, and no paired input and output sentences are available. The core of our framework is a latent decomposition scheme learned via adversarial training. We decompose the latent representation of a sentence into two parts where one encodes the style of a sentence, while the other encodes the style-independent content of the sentence. In the test time, for changing the style of an input sentence, we first extract its content code. We then sample a sentence from the training dataset of the target style corpus and extract its style code. The two codes are combined to generate an output sentence, which would carry the same content but in the target style. As sampling a different style sentence, we have a different style code and have a different style transfer output. We conduct experiments with comparison to several state-of-the-art approaches on multiple public datasets, including Yelp~\cite{yelp-challenge} and Amazon~\cite{he2016ups}. The results, evaluated using various performance metrics, including content preservation, style accuracy, output diversity, and user preference, show that the model trained with our framework performs consistently better than the competing approaches.

\section{Methodology}

Let $X_1$ and $X_2$ be two spaces of sentences of two different language styles. Let $Z_1$ and $Z_2$ be their corresponding latent spaces. We further assume $Z_1$ and $Z_2$ can be decomposed into two latent spaces $Z_1=S_1 \times C_1$ and $Z_2=S_2 \times C_2$ where $S_1$ and $S_2$ are the latent spaces that control the style variations in $X_1$ and $X_2$ and $C_1$ and $C_2$ are the latent spaces that control the style-independent content information. Since $C_1$ and $C_2$ are style-independent content representation, we have $C \equiv C_1 \equiv C_2$. For example, $X_1$ and $X_2$ may denote the spaces of negative and positive product reviews where the elements in $C$ encode the product and its features reviewed in a sentence, the elements in $S_1$ represent variations in negative styles such as the degree of preferences and the exact phrasing, and the elements in $S_2$ represent the corresponding variations in positive styles. The above modeling implies
\begin{enumerate}[noitemsep,nolistsep,leftmargin=*]
	\item A sentence $x_1 \in X_1$ can be decomposed to a content code $c_1 \in C$ and a style code $s_1 \in S_1$.
	\item A sentence $x_1 \in X_1$ can be reconstructed by fusing its content code $c_1$ and its style code $s_1$.
	\item To transfer a sentence in $X_1$ to a corresponding sentence in $X_2$, one can simply fuse the content code $c_1$ with a style code $s_2$ where $s_2 \in S_2$.
\end{enumerate}
Figure~\ref{fig:problem} provides a visualization of the modeling. 

Under this formulation, the text style transfer mechanism is given by a conditional distribution $p(x_{1 \rightarrow 2} | x_1)$, where $x_{1 \rightarrow 2}$ is the sentence generated by transferring sentence $x_1$ to the target domain $X_2$. Note that existing works~\cite{fu2018style,shen2017style} formulate the text style transfer mechanism to be a one-to-one mapping that converts an input sentence to only a single corresponding output sentence. That is $p(x_{1 \rightarrow 2} | x_1)=\delta (x_1)$ where $\delta$ is the Dirac delta function. As a results, they can not be used to generate multiple style transfer outputs for an input sentence.

\begin{figure}[tb]
	\centering
	\ \ \ \ \ \ \  \includegraphics[height=.2\textwidth]{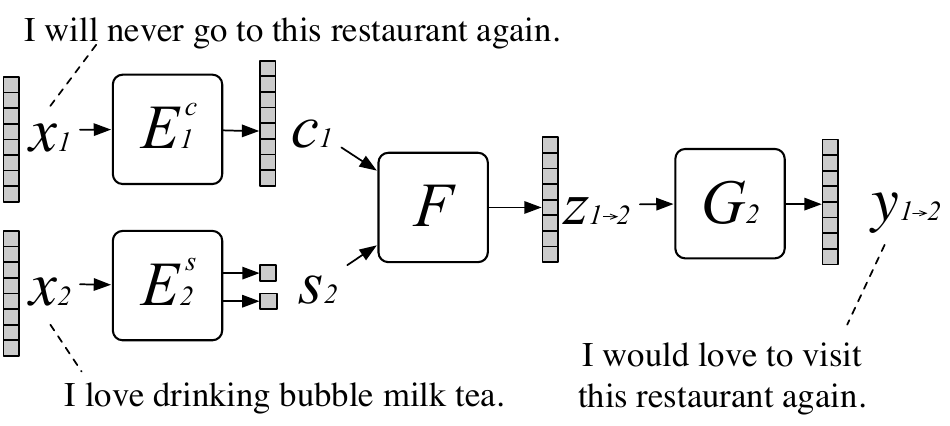}
	\centering
	\includegraphics[height=.2\textwidth]{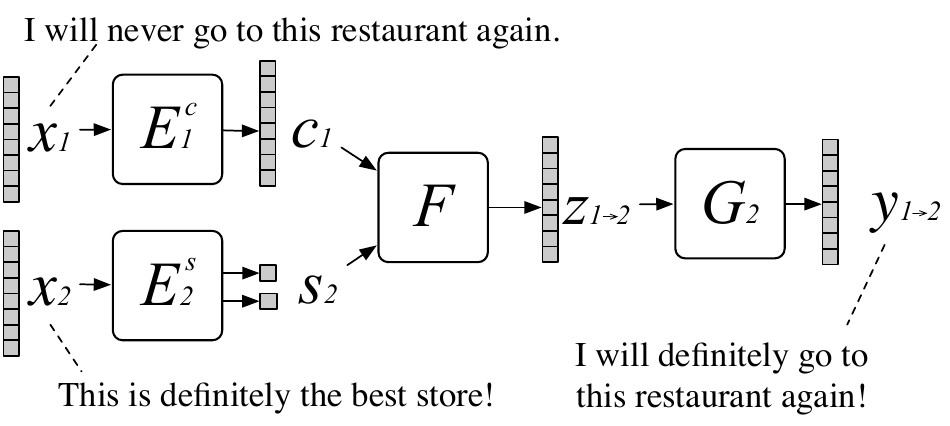}
	\caption{Overview of the proposed one-to-many style transfer approach. 
	We show an example of transferring a negative restaurant review sentence $x_1$ to multiple different positive ones $y_{1\rightarrow2}$.
	To transfer the sentence, we first randomly sample a sentence $x_2$ from the space of positive reviews $X_2$ and extract its style code $s_2$ using $E_2^s$. We then compute $z_{1\rightarrow 2}$ by combining $c_1$ with $s_2$ and convert it to the transfer output $y_{1\rightarrow 2}$ using $G_2$. We note that by sampling a different $x_2$ and hence a different $s_2$, we have a different style transfer output $y_{1\rightarrow2}$.}
	\label{fig:network}
	\vspace{-2mm}
\end{figure}

\noindent{\bf One-to-Many Style Transfer.} To model the transfer function, we use a framework consists of a set of networks as visualized in Figure~\ref{fig:network}. It has a content encoder $E_i^c$, a style encoder $E_i^s$, and a decoder $G_i$ for each domain $X_i$. In the following, we will explain the framework in details using the task of transferring from $X_1$ to $X_2$. The task of transferring from $X_2$ to $X_1$ follows the same pattern.

The content encoder $E_1^c$ takes the sequence $x_1 = \{x_1^1, x_1^2, \dots, x_1^{m_{1}(x_1)}\}$ of $m_1(x_1)$ elements as input and computes a content code $c_1 \equiv \{c_1^1, c_1^2, \dots, c_1^{m_{1}(x_1)}\} = E_1^c(x_1)$, which is a sequence of vectors describing the sentence's style-independent content. The style encoder $E_2^S$ converts $x_2$ to a style code $s_2\equiv(s_{2,\mu}, s_{2,\sigma})=E_2^s(x_2)$, which is a pair of vectors. Note that we will use $s_{2,\mu}$ and $s_{2,\sigma}$ as the new mean and standard deviation of the feature activation of the input $x_1$ for the style transfer task of converting a sentence in $X_1$ to a corresponding sentence in $X_2$. Specifically, we combine the content code $c_1$ and the style code $s_2$ using a composition function $F$, which will be discussed momentarily, to obtain $z_{1\rightarrow 2}=\{z_{1\rightarrow 2}^1, z_{1\rightarrow 2}^2, \dots, z_{1\rightarrow 2}^{m_{1}({x_{1}})}\}$. Then, we use the decoder $G_2$ to map the representation $z_{1\rightarrow 2}$ to the output sequence $y_{1\rightarrow 2}$. Note that $s_2$ is extracted from a randomly sampled $x_2\in X_2$, and by sampling a different sentence, say $x_2^{\prime}\in X_2$ where $x_2^{\prime} \neq x_2$, we have $s_2^{\prime}\neq s_2$ and hence a different style transfer output. By treating style variations as sample-able quantities, we achieve one-to-many style transfer output capability.

The combination function is given by 
\begin{equation}
F(c_i^k, s_j) = s_{j,\sigma} \otimes (c_i^k - \mu(c_i)) \oslash \sigma(c_i) + s_{j,\mu},
\end{equation}
where $\otimes$ denotes element-wise product, $\oslash$ denotes element-wise division, $\mu(\cdot)$ and $\sigma(\cdot)$ indicate the operation of computing mean and standard derivation for the content latent code by treating each vector in $c_i$ as an independent realization of a random variable. In other words, the latent representation $z^{k}_{i\rightarrow j}=F(c^k_i, s_j)$ is constructed by first normalizing the content code $c_i$ in the latent space and then applying the non-linear transformation whose parameters are provided from a sentence of target style. Since $F$ contains no learnable parameters, we consider $F$ as part of the decoder. This design draws inspirations from image style transfer works~\cite{huang2017arbitrary,dumoulin2016learned}, which show that image style transfer can be achieved by controlling the mean and variance of the feature activations in the neural networks. We hypothesize this is the same case for the text style transfer task and apply it to achieve the one-to-many style transfer capability.

\noindent{\bf Network Design.} We realize the content encoder $E_i^c$ using a convolutional network. To ensure the length of the output sequence $c$ is equal to the length of the input sentence, we pad the input by $m-1$ zero vectors on both left and right side, where $m$ is the length of the input sequence as discussed in~\cite{gehring2017convolutional}. For the convolution operation, we do not include any stride convolution. We also realize the style encoder $E_i^s$ using a convolutional network. To extract the style code, after several convolution layers, we apply global average pooling and then project the results to $s_{i,\mu}$ and $s_{i,\sigma}$ using a two-layer multi-layer perceptron. We apply the log-exponential nonlinearity to compute $s_{i,\sigma}$ to ensure the outputs are strictly positive, required for modeling the deviations. The decoder $G_i$ is realized using a convolutional network with an attention mechanism followed by a convolutional sequence-to-sequence network (ConvS2S)~\cite{gehring2017convolutional}. 
We realized our method based on ConvS2S, but it can be extended to work with transformer models~\cite{vaswani2017attention,devlin2018bert,radford2019language}. Further details are given in the appendix.

\subsection{Learning Objective}

We train our one-to-many text style transfer model by minimizing multiple loss terms.

{\flushleft\textbf{Reconstruction loss.} We use reconstruction loss to regularize the text style transfer learning. Specifically, we assume the pair of content encoder $E_i^c$ and style encoder $E_i^s$ and the decoder $G_i$ form an auto-encoder. We train them by minimizing the negative log likelihood of the training corpus:}
\begin{equation}
\mathcal{L}_{rec}^i = \mathbb{E}_{x_i} [-\log P(y_i^k|x_i^k;\theta_{E_i^c},\theta_{E_i^s},\theta_{G_i})]
\end{equation}
where $\theta_{E_i^c}$, $\theta_{E_i^s}$ and $\theta_{G_i}$ denote the parameters of $E_i^c$, $E_i^s$, and $G_i$ respectively. 

For each training sentence, $G_i$ synthesizes the output sequence by predicting the most possible token $y^t$ based on the latent representation $z_i\equiv\{z_i^1,z_i^2,...,z_i^m\}$ and the previous output predictions $\{y^1, y^2, \dots, y^{t-1}\}$, so that the probability of a sentence can be calculated by
\begin{align}
P(y|x;&\theta_{E_i^c},\theta_{E_i^s},\theta_{G_i}) = \nonumber\\
&\prod_{t=1}^{T} p(y^t|z_i,y^1, y^2, \dots, y^{t-1};\theta_{G_i}),
\end{align}
where $t$ denotes the token index and $T$ is the sentence length. 
Following~\cite{gehring2017convolutional}, the probability of a token is computed by the linear projection of the decoder output using softmax.

\begin{figure*}
\centering
\includegraphics[width=.75\textwidth]{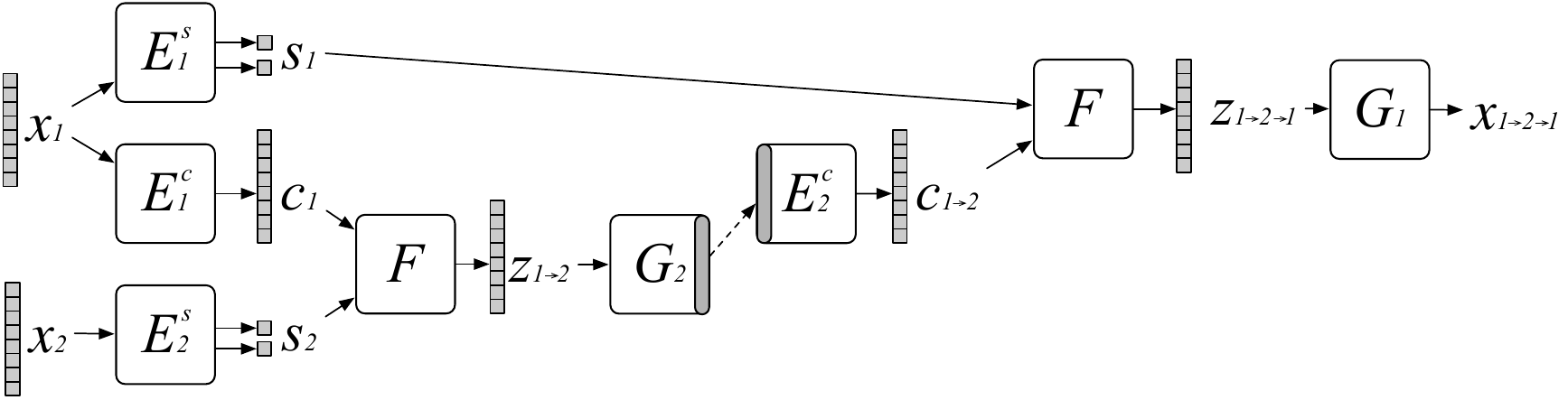}
\caption{Illustration of the back-translation loss. We transfer $x_1$ to the domain of $X_2$ and then transfer it back to the domain of $X_1$ using its original style code $s_1$. The resultant sentence $x_{1\rightarrow2\rightarrow1}$ should be as similar as possible to $x_1$ if the content code is preserved across transfer. To tackle the non-differentiable of the sentence decoding mechanism (beam search), we replace the hard decoding of $x_{1\rightarrow2}$ by a learned non-linear projections between the decoder $G_2$ and the content encoder $E_2^c$.}\label{fig:cyc}
\end{figure*}

{\flushleft\textbf{Back-translation loss.} Inspired by recent studies~\cite{prabhumoye2018style,sennrich2015improving,brislin1970back} that show that back-translation loss, which is closely related to the cycle-consistency loss~\cite{CycleGAN2017} used in computer vision, is helpful for preserving the content of the input, we adopt a back-translation loss to regularize the learning. To achieve the goal, as shown in Figure~\ref{fig:cyc}, we transfer the input $x_1$ to the other style domain $X_2$. We then transfer it back to the original domain $X_1$ by using its original style code $s_1$. By doing so, the resulting sentence $x_{1\rightarrow2\rightarrow1}$ should be as similar as possible to the original input $x_1$. In other words, we minimize the discrepancy between $x_{1}$ and $x_{1\rightarrow2\rightarrow1}$ given by}
\begin{equation}
\begin{aligned}\label{eqn:cyc-loss}
&\mathcal{L}_{back}^1 = \mathbb{E}_{x_1,x_2}[-\log P(y_1^k|x_{1\rightarrow2\rightarrow1}^k;\underline{\theta})]
\end{aligned}
\end{equation}
where $\underline{\theta}=\{\theta_{E_1^c},\theta_{E_1^s},\theta_{G_1},\theta_{E_2},\theta_{E_2^s},\theta_{G_2}\}$. We also define $\mathcal{L}_{back}^2$ in a similar way.

To avoid the non-differentiability of the beam search~\cite{och2004alignment,sutskever2014sequence}, we substitute the hard decoding of $x_{1\rightarrow2}$ by using a set of differentiable non-linear transformations between the decoder $G_2$ and the content encoder $E_1^c$ when minimizing the back-translation loss. The non-linear transformations project the feature activation of the second last layer of the decoder $G_2$ to the second layer of the content encoder $E_1^c$. These non-linear projections are learned by the multilayer perceptron (MLP), which are trained jointly with the text style transfer task. We also apply the same mechanism to compute $x_{2\rightarrow1}$. This way, our model can be trained purely using back-propagation.

To ensure the MLP correctly project the feature activation to the second layer of $E_2^c$, we enforce the output of the MLP to be as similar as possible to the feature activation of the second layer of $E_1^c$. This is based on the idea that $x_1$ and $x_{1\rightarrow2}$ should have the same content code across transfer, and their feature activation in the content encoder should also be the same. Accordingly, we apply Mean Square Error (MSE) loss function to achieve this objective:
\begin{equation}%
\begin{aligned}%
\label{eqn:mse-loss}%
\mathcal{L}_{mse}^{1} = \mathbb{E}_{x_1,x_2}[||E_2^{c,h}(x_{1\rightarrow 2})-E_1^{c,h}(x_1)||_2^2]
\end{aligned}
\end{equation}
where $E_1^{c,h}$ and $E_2^{c,h}$ denote the function for computing feature activation of the second layer of $E_1^c$ and $E_2^c$, respectively. The loss $\mathcal{L}_{mse}^{2}$ for the other domain is defined in a similar way.

{\flushleft\textbf{Style classification loss.} During learning, we enforce a style classification loss on the style code $s_i = E_i^s(x_i)$ with the standard cross-entropy loss $\mathcal{L}^i_{cls}$. This encourages the style code $s_i$ to capture the stylistic properties of the input sentences.}

{\flushleft\textbf{Adversarial loss.} We use GANs~\cite{goodfellow2014generative} for matching the distribution of the input latent code to the decoder from the reconstruction streams to the distribution of the input latent code to the decoder from the translation stream. That is (1) we match the distribution of $z_{1\rightarrow 2}$ to the distribution of $z_{2}$, and (2) we match the distribution of $z_{2\rightarrow 1}$ to the distribution of $z_{1}$. This way we ensure distribution of the transfer outputs matches distribution of the target style sentences since they use the same decoder. As we apply adversarial training to the latent representation, we also avoid dealing with the non-differentiability of beam search.}	
The adversarial loss for the second domain is given by
\begin{equation}%
\begin{aligned}\label{eqn:gan-loss}%
\mathcal{L}_{adv}^{2} & = \mathbb{E}_{x_1,x_2}\left[ \log(1-D_2(z_{1\rightarrow 2}))\right] \\
& + \mathbb{E}_{x_2} \left[\log(D_2(z_2)) \right],
\end{aligned}
\end{equation}
where $D_2$ is the discriminator which aims at distinguishing the latent representation of the sentence $z_{1\rightarrow2}$ from $z_2=\mathcal{C}_z(c_2,s_2)$. The adversarial loss $\mathcal{L}_{adv}^{1}$ is defined in a similar manner.

{\flushleft\textbf{Overall learning objective.} We then learn a one-to-many text style transfer model by solving}%
\begin{equation}%
\begin{aligned}\label{eqn:gan-loss}%
\min_{E_1,E_2,G_1,G_2}  \max_{D_1,D_2} \sum_{i=1}^{2} \big{(} \mathcal{L}^i_{rec} + \mathcal{L}^i_{back} + \mathcal{L}^i_{mse} \\+ \mathcal{L}^i_{cls} + \mathcal{L}^i_{adv} \big{)}.
\end{aligned}
\end{equation}

\section{Experiments}

In the following, we first introduce the datasets and evaluation metrics and then present the experiment results with comparison to the competing methods.

\noindent\textbf{Datasets.} We use the following datasets.
\begin{itemize}[leftmargin=*,noitemsep,topsep=0pt]
\item \textbf{Amazon product reviews (Amazon)} \cite{he2016ups} contains $277,228$ positive and $277,769$ negative review sentences for training, and $500$ positive and $500$ negative review sentences for testing. The length of a sentence ranges from $8$ to $25$ words. We use this dataset for converting a negative product review to a positive one, and vice versa. Our evaluation follows the protocol described in \citet{li2018delete}.
\item \textbf{Yelp restaurant reviews (Yelp)}~\cite{yelp-challenge} contains a training set of $267,314$ positive and $176,787$ negative sentences, and a test set of $76,392$ positive and $50,278$ negative testing sentences. The length of a sentence ranges from $1$ to $15$ words. We use this dataset for converting a negative restaurant review to a positive one, and vice versa. We use two evaluation settings: Yelp500 and Yelp25000. Yelp500 is proposed by~\cite{li2018delete}, which includes randomly sampled $500$ positive and $500$ negative sentences from the test set, while Yelp25000 includes randomly sampled $25000$ positive and $25000$ negative sentences from the test set.
\end{itemize}

\noindent\textbf{Evaluation metrics.} We evaluate a text style transfer model on several aspects. Firstly, the transfer output should carry the target style (style score). Secondly, the style-independent content should be preserved (content preservation score). We also measure the diversity of the style transfer outputs for an input sentence (diversity score).

\begin{itemize}[leftmargin=*,noitemsep,topsep=0pt]
\item \textbf{Style score.} We use a classifier to evaluate the fidelity of the style transfer results\cite{fu2018style,shen2017style}. Specifically, we apply the Byte-mLSTM~\cite{radford2017learning} to classify the output sentence generated by a text style transfer model. As transferring a negative sentence to a positive one, we expect a good transfer model should be able to generate a sentence that is classified positive by the classifier. The overall style transfer performance of a model is then given by the average accuracy on the test set measured by the classifier. 
\item\textbf{Content score.} We build a style-independent distance metric that can quantify content similarity between two sentences, by comparing embeddings of the sentences after removing their style words. Specifically, we compute embedding of each non-style word in the sentence using the word2vec~\cite{mikolov2013distributed}. Next, we compute the average embedding, which serves as the content representation of the sentence. The content similarity between two sentences is given by the cosine distance of their average embeddings. We compute the relative n-gram frequency to determine which word is a style word based on the observation that the language style is largely encoded in the n-gram distribution~\cite{xu2012paraphrasing}. This is in spirit similar to the term frequency-inverse document frequency analysis~\cite{sparck1972statistical}. Let $D_1$ and $D_2$ be the n-gram frequencies of two corpora of different styles. The style magnitude of an n-gram $u$ in style domain $i$ is given by \begin{equation}
s_i(u) = \frac{D_i(u)+\lambda}{\sum_{j\neq i} D_j(u)+\lambda}
\end{equation}
where $\lambda$ is a small constant. We use $1$-gram. A word is considered a style word if 
$\min_{k\in\{i,j\}} s_{k}(u)$ is greater than a threshold.
\item\textbf{Diversity score.} To quantify the diversity of the style transfer outputs, we resort to the self-BLEU score proposed by~\citet{zhu2018texygen}. Given an input sentence, we apply the style transfer model 5 times to obtain 5 outputs. We then compute self-BLEU scores between any two generated sentences (10 pairs). We apply this procedure to all the sentences in the test set and compute the average self-BLEU score $v$. After that, we define the diversity score as $100-v$. A model with a higher diversity score means that the model is better in generating diverse outputs. In the experiments, we denote Diversity-$K$ as the diversity score computed by using self-BLEU-$K$. 
\end{itemize}

\noindent\textbf{Implementation.} We use the convolutional sequence-to-sequence model~\cite{gehring2017convolutional}. Our content and style encoder consist of $3$ convolution layers, respectively. The decoder has $4$ convolution layers. The content and style codes are $256$ dimensional. We use the \texttt{pytorch}~\cite{paszke2017automatic} and \texttt{fairseq}~\cite{ott2019fairseq} libraries and train our model using a single GeForce GTX 1080 Ti GPU. We use the SGD algorithm with the learning rate set to $0.1$. Once the content and style scores converge, we reduce the learning rate by an order of magnitude after every epoch until it reaches $0.0001$. Detail model parameters are given in the appendix.

\noindent\textbf{Baselines.} We compare the proposed approach to the following competing methods.
\begin{itemize}[leftmargin=*,noitemsep,topsep=0pt]
\item\textbf{CAE}~\cite{shen2017style} is based on auto-encoder and is trained using a GAN framework. It assumes a shared content latent space between different domains and computes the content code by using a content encoder. The output is generated with a pre-defined binary style code.
\item\textbf{MD}~\cite{fu2018style} extends the CAE to work with multiple style-specific decoders. It learns style-independent representation by adversarial training and generates output sentences by using style-specific decoders.
\item\textbf{BTS}~\cite{prabhumoye2018style} learns style-independent representations by using back-translation techniques. BTS assumes the latent representation of the sentence preserves the meaning after machine translation.
\item\textbf{DR}~\cite{li2018delete} employs retrieval techniques to find similar sentences with desired style. They use neural networks to fuse the input and the retrieved sentences for generating the output.
\item\textbf{CopyPast} simply uses the input as the output, which serves as a reference for evaluation.
\end{itemize}

\begin{table}[t]
	\centering
	\small
		\begin{tabular}{lccc}
			\toprule
			\textbf{Amazon}  & Diversity-$4$ &  Diversity-$3$ & Diversity-$2$ \\
			\midrule
			CAE & $2.60$ & $2.15$ & $1.64$ \\
			\midrule
			CAE+noise & $33.01$ & $29.33$ & $24.66$ \\ 
			BTS+noise & $39.22$ & $35.46$ & $30.48$ \\ 
			\midrule
			Ours & $\textbf{46.31}$ & $\textbf{41.69}$ & $\textbf{36.01}$ \\
			\bottomrule
		\end{tabular}
	\small
		\begin{tabular}{lccc}\\
			\toprule
			\textbf{Yelp}  & Diversity-$4$ &  Diversity-$3$ & Diversity-$2$ \\
			\midrule
			CAE & $1.03$ & $0.80$ & $0.60$ \\
			\midrule
			CAE+noise & $16.91$ & $14.63$ & $11.73$ \\ 
			BTS+noise & $48.36$ & $43.69$ & $37.38$ \\ 
			\midrule
			Ours & $\textbf{58.29}$ & $\textbf{50.90}$ & $\textbf{42.34}$ \\
			\bottomrule
		\end{tabular}
	\caption{One-to-many text style transfer results.}
	\label{tbl:yelp-diverse}
	\label{tbl:amazon-diverse}
\center
	\small
\begin{tabular}{lccc}
    \toprule
	Method  & Diversity& Fluency & Overall\\
	\midrule
	CAE+noise & 13.13 & 11.62 & 12.12\\
	No Pref. & 35.35 & 16.16 & 36.87\\
  	Ours & \textbf{51.52} & \textbf{72.22} & \textbf{51.01}\\  
	\midrule
    BTS+noise & 13.13 & 11.11 & 16.16\\
    No Pref. & 42.93 & 22.22 & 40.40\\
    Ours & \textbf{43.94} & \textbf{66.67} & \textbf{43.43}\\
	\bottomrule
	\vspace{0.08mm}
\end{tabular}

	\caption{User study results on one-to-many text style transfer. The numbers are the user preference score of competing methods.}
	\label{tbl:user2}
\end{table}

\begin{table}[t]
\centering
\fbox{
\footnotesize
\begin{minipage}{24em}
\raggedright
\textbf{Input:} I will never go to this restaurant again.\\
\textbf{Output A:} I will {\color{blue}{definitely}} go to this restaurant again.\\
\textbf{Output B:} I will {\color{blue}{continue}} go to this restaurant again.\\
\textbf{Output C:} I will {\color{blue}{definitely}} go to this {\color{blue}{place}} again.\\
\rule[0.3\baselineskip]{\textwidth}{0.05pt}
\textbf{Input:} It was just a crappy experience over all.\\
\textbf{Output A:} It was just a {\color{blue}{wonderful}} experience {\color{blue}{at}} all.\\
\textbf{Output B:} {\color{blue}{Great place}} just a {\color{blue}{full}} experience over all.\\
\textbf{Output C:} It was {\color{blue}{such}} a {\color{blue}{good}} experience {\color{blue}{as}} all.\\
\end{minipage}
}
\fbox{
\footnotesize
\begin{minipage}{24em}
\raggedright
\textbf{Lyrics input:} My friends they told me you change like the weather; From one love to another you would go; But when I first met you your love was like the summer; Love I never dreamed of turning cold\\
\textbf{Romantic style:} My friends they told me you change like the {\color{blue}{light}}; From one love to another you would go; But when I first met you your love was like the {\color{blue}{sun}}; Love I never dreamed of turning cold\\
\textbf{Romantic style:} My {\color{blue}{lips}} they told me you change like the {\color{blue}{light}}; From one love to {\color{blue}{find}} you would go; But when I {\color{blue}{am}} you your love was like the {\color{blue}{mountain}}; Love I never {\color{blue}{wanted}} of {\color{blue}{me before}}\\
\end{minipage}
}
\caption{One-to-many style transfer results computed by the proposed algorithm.}
\label{tbl:qual_sent}
\end{table}

\subsection{Results on One-to-Many Style Transfer}

Our model can generate different text style transfer outputs for an input sentence. To generate multiple outputs for an input, we randomly sample a style code from the target style training dataset during testing. Since the \textbf{CAE}~\cite{shen2017style} and \textbf{BTS}~\cite{prabhumoye2018style} are not designed for the one-to-many style transfer, we extend their methods to achieve this capability by injecting random noise, termed \textbf{CAE+noise} and \textbf{BTS+noise}. Specifically, we add random Gaussian noise to the latent code of their models during training, which is based on the intuition that the randomness would result in different activations in the networks, leading to different outputs. Table~\ref{tbl:yelp-diverse} shows the average diversity scores achieved by the competing methods over $5$ runs. We find that our method performs favorably against others. 

\noindent\textbf{User Study.} We conduct a user study to evaluate one-to-many style transfer performance using the Amazon Mechanical Turk (AMT) platform. We set up the pairwise comparison following~\citet{prabhumoye2018style}. Given an input sentence and two sets of model-generated sentences (5 sentences per set), the workers are asked to choose which set has more diverse sentences with the same meaning, and which set provides more desirable sentences considering both content preservation and style transfer. These are denoted as \textit{Diversity}, and \textit{Overall} in Table~\ref{tbl:user2}. The workers are also asked to compare the transfer quality in terms of grammatically and fluency, which is denoted as \textit{Fluency}. For each comparison, a third option \textit{No Preference} is given for cases that both are equally good or bad. 

We randomly sampled $250$ sentences from Yelp500 test set for the user study. Each comparison is evaluated by at least three different workers. We received more than $3,600$ responses from the AMT, and the results are summarized in Table~\ref{tbl:user2}. Our method outperforms the competing methods by a large margin in terms of diversity, fluency, and overall quality. In the appendix, we present further details of the comparisons with different variants of \textbf{CAE+noise} and \textbf{BTS+noise}. Our method achieves significantly better performance. Table~\ref{tbl:qual_sent} shows the qualitative results of the proposed method. Our proposed method generates multiple different style transfer outputs for restaurant reviews and lyrics\footnote{We use the country song lyrics and romance novel collections, which are available in the Stylish descriptions dataset~\cite{chen2019unsupervised}.}.

\subsection{More Results and Ablation Study}

\begin{figure*}[t]
	\centering
	\subfloat[Amazon]{\label{fig:amazon}{\includegraphics[width=.33\textwidth]{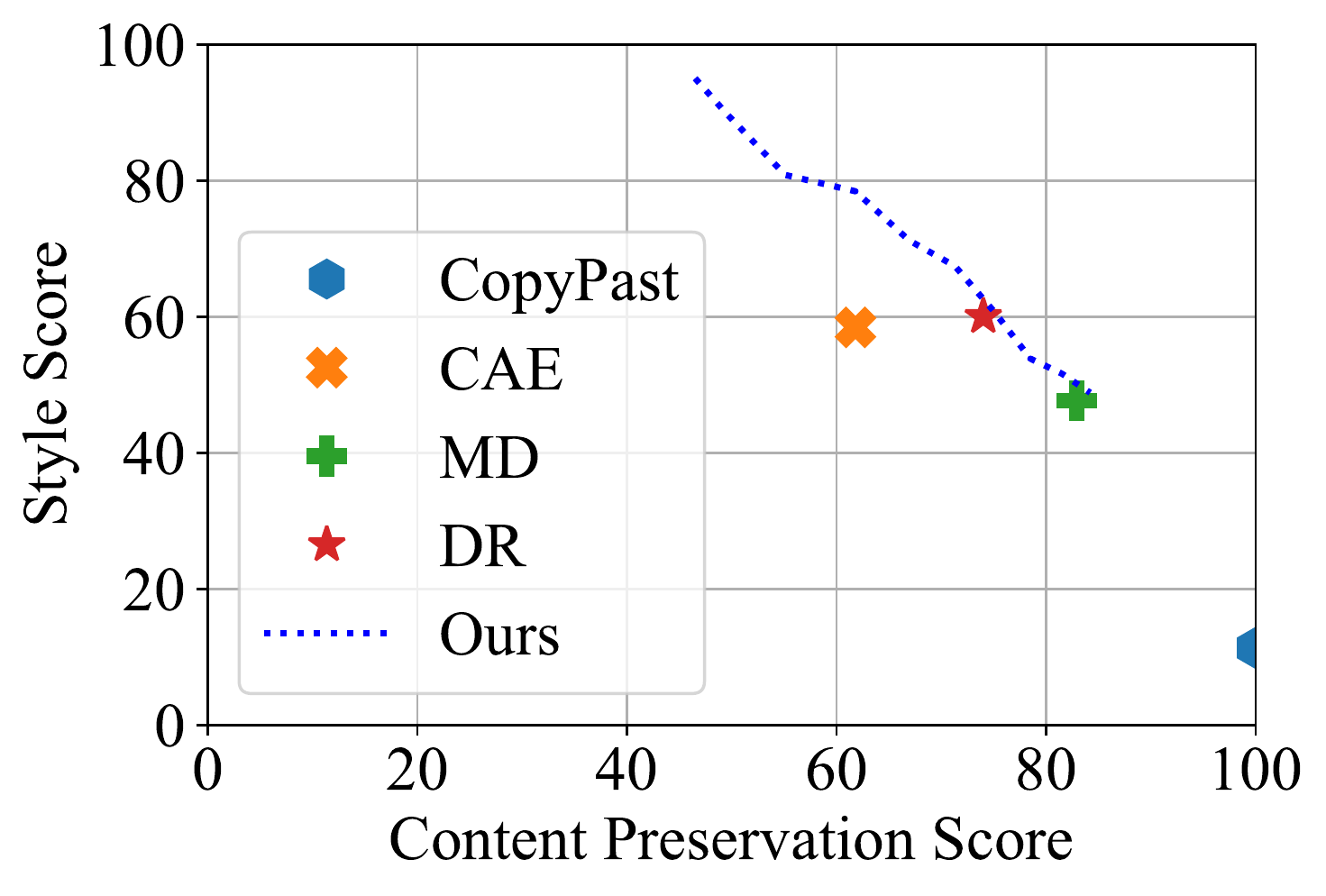}}}
	\subfloat[Yelp500]{\label{fig:smallyelp}{\includegraphics[width=.33\textwidth]{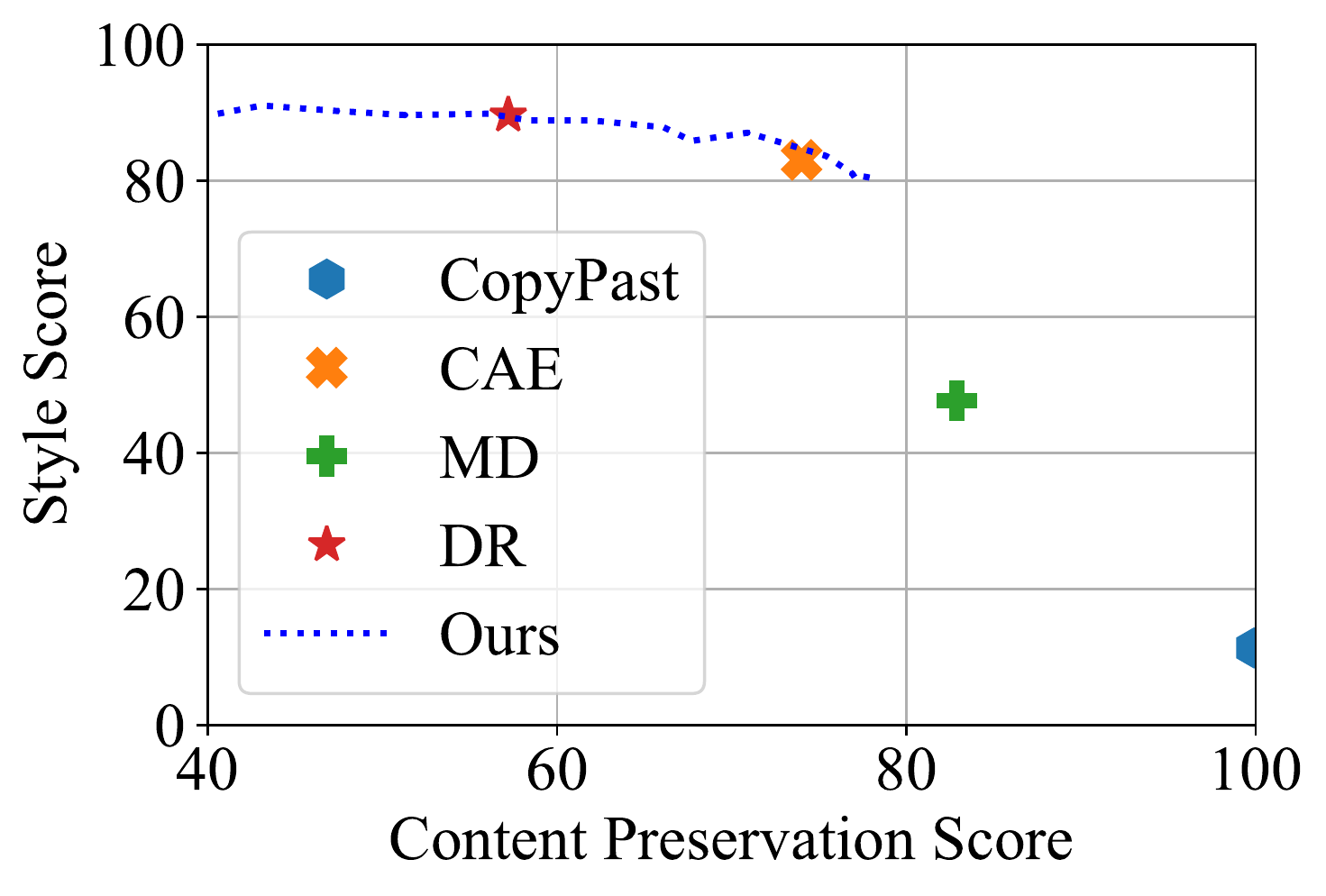}}}
	\subfloat[Yelp25000]{\label{fig:yelp}{\includegraphics[width=.33\textwidth]{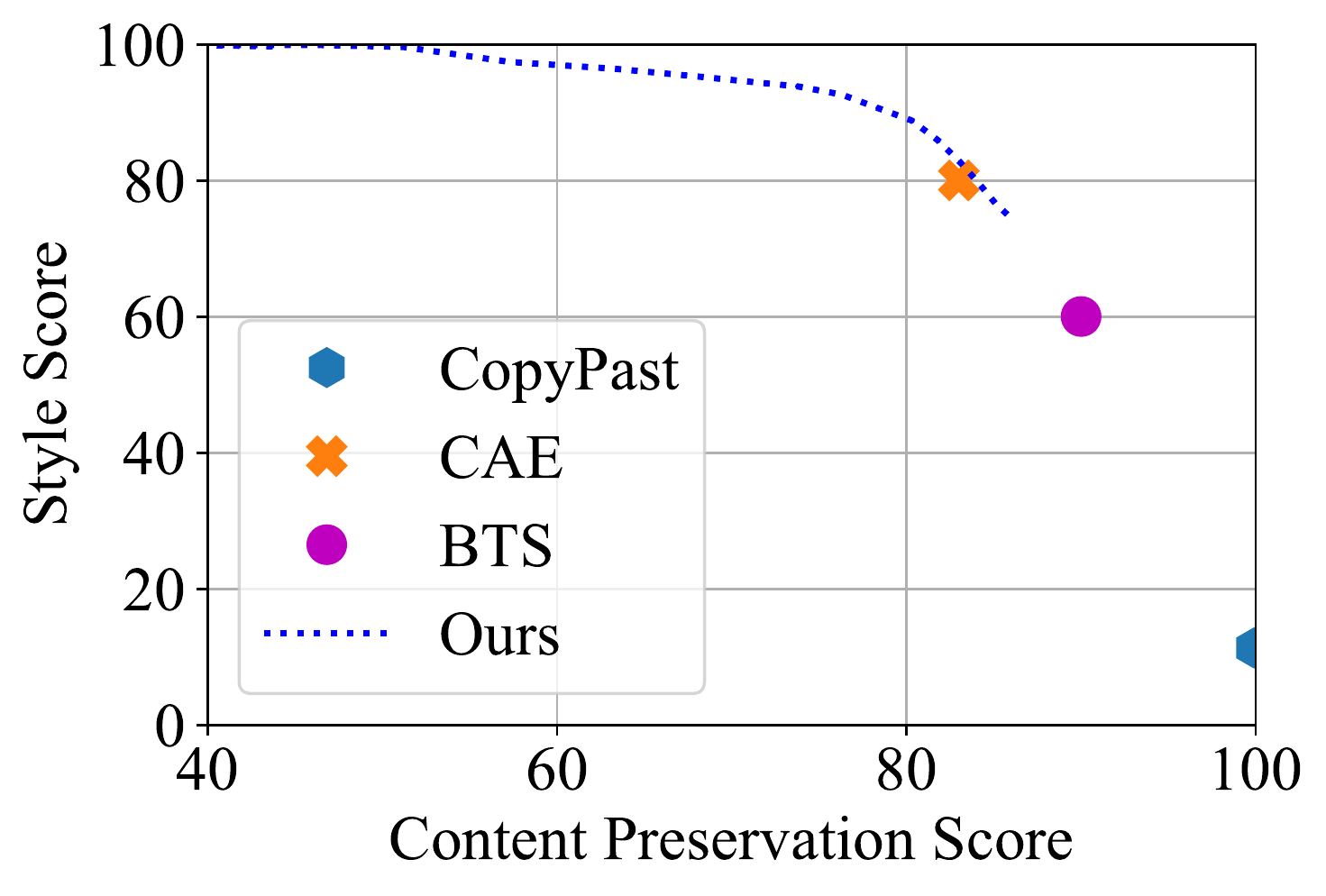}}}\\
	\caption{Comparison to different style transfer algorithms on output quality.}
	\label{fig:comparison}
\end{figure*}

In addition to generating multiple style transfer outputs, our model can also generate high-quality style transfer outputs. In Figure~\ref{fig:comparison}, we compare the quality of our style transfer outputs with those from the competing methods. We show the performance of our model using the style--content curve where each point in the curve is the achieved style score and the content score at different training iterations. In Figure~\ref{fig:amazon}, given a fixed content preservation score, our method achieves a better style score on Amazon dataset. Similarly, given a fixed style score, our model achieves a better content preservation score. The results on Yelp500 and Yelp25000 datasets also demonstrate a similar trend as shown in Figure~\ref{fig:smallyelp} and Figure~\ref{fig:yelp}, respectively. 

\begin{figure}
        \centering
		\includegraphics[height=0.095\textheight]{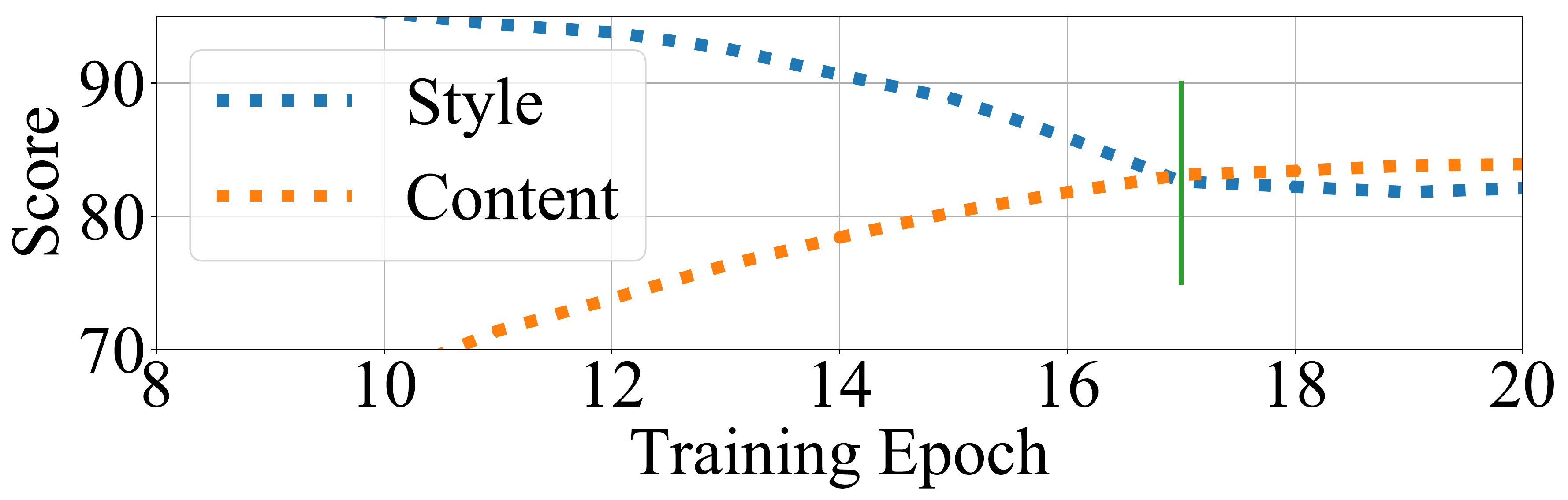}
		\caption{Style--content trade-off curves. The vertical line indicates the iteration at which the learning rate is decreased.}
		\label{fig:tradeoff}
\end{figure}

The style--content curve also depicts the behavior of the proposed model during the entire learning process. As visualized in Figure~\ref{fig:tradeoff}, we find that our model achieves a high style score but a low content score in the early training stage. With more iterations, our model improves the content score with the expense of a reduced style score. To strike a balance between the two scores, we decrease the learning rate when the model reaches a similar number for the two scores.

\begin{table}[t]
\center
	\small
\begin{tabular}{lcccc}
    \toprule
	Method  & Style & Content & Fluency & Overall\\
	\midrule
	CAE & 30.56 & 36.81 & 23.26 & 30.56\\
	No Pref. & 31.60 & \textbf{39.93} & \textbf{51.74} & \textbf{37.50}\\
    Ours & \textbf{37.85} & 23.26 & 25.00 & 31.94\\
	\midrule
	MD & 29.26 & 27.56 & 27.35 & 28.41\\
	No Pref. & 21.88 & \textbf{52.84} & \textbf{47.01} & 29.83\\
    Ours & \textbf{48.86} & 19.60 & 25.64 & \textbf{41.76}\\
	\midrule
	BTS & 30.66 & \textbf{40.88} & 16.79 & 30.29\\
	No Pref. & 31.75 & 22.99 & \textbf{56.93} & 32.12\\
    Ours & \textbf{37.59} & 36.13 & 26.28 & \textbf{37.59}\\
	\midrule
	DR & 26.30 & 18.09 & 22.61 & 24.96\\
	No Pref. & 11.89 & \textbf{69.01} & \textbf{57.96} & 21.61\\
    Ours & \textbf{61.81} & 12.90 & 19.43 & \textbf{53.43}\\
	\bottomrule
\end{tabular}
	\caption{User study results. The numbers are the user preference scores of the competing methods.}
	\label{tbl:user}
\end{table}

\begin{table}[t]
	\small
		\centering
		\begin{tabular}{lcc}
			\toprule
			Model  & Style Score & Content Score\\
			\midrule
			\textit{sharing-decoder}  & $63.75$ & $42.54$\\
			\textit{sharing-encoders} & $81.41$ & $81.48$\\
			\textit{full}             & $\textbf{82.64}$ & $\textbf{83.11}$\\
			\bottomrule
		\end{tabular}
		\caption{Comparison of different design choices of the proposed framework.}
		\label{tbl:model_design}
\end{table}

\noindent\textbf{User Study.} We also conduct a user study on the transfer output quality. Given an input sentence with two generated style transferred sentences from two different models\footnote{The sentences generated by other methods have been made publicly available by~\cite{li2018delete}.}, workers are asked to compare the transferred quality of the two generated sentences in terms of content preservation, style transfer, fluency, and overall performance, respectively. We received more than $2500$ responses from AMT platform, and the results are summarized in Table~\ref{tbl:user}. We observe \textit{No Preference} was chosen more often than others, which shows exiting methods may not fully satisfy human expectation. However, our method achieves comparable or better performance than the prior works.

\noindent\textbf{Ablation Study.} We conduct a study where we consider three different designs of the proposed models. (1) \textit{full}: This is the full version of the proposed model; (2) \textit{sharing-encoders}: In this case, we have a content encoder and a style encoder that are shared by the two domains; (3) \textit{sharing-decoder}: In this case, we have a decoder that is shared by the two domains. Through this study, we aim for studying if regularization via weight-sharing is beneficial to our approach.

Table~\ref{tbl:model_design} shows the comparison of our method using different designs. The \textit{sharing-encoders} baseline performs much better than the \textit{sharing-decoder} baseline, and our \textit{full} method performs the best. The results show that the style-specific decoder is more effective for generating target-style outputs. On the other hand, the style-specific encoder extracts more domain-specific style codes from the inputs. Weight-sharing schemes do not lead to a better performance.

\noindent\textbf{Impact of the loss terms.} In the appendix, we present an ablation study on the loss terms, which shows that all the terms in our objective function are important.

\section{Related Works}

\noindent{\bf Language modeling} is a core problem in natural language processing. It has a wide range of applications including machine translation~\cite{johnson2017google,wu2016google}, image captioning~\cite{vinyals2015show}, and dialogue systems~\cite{li2016persona,li2016deep}. Recent studies~\cite{devlin2018bert,gehring2017convolutional,graves2013generating,johnson2017google,radford2019language,wu2016google} proposed to train deep neural networks using maximum-likelihood estimation (MLE) for computing the lexical translation probabilities in parallel corpus. Though effective, acquiring parallel corpus is difficult for many language tasks. 

\noindent{\bf Text style transfer} has a longstanding history~\cite{ker1992style}. Early studies utilize strongly supervision on parallel corpus~\cite{rao2018dear,xu2017shakespeare,xu2012paraphrasing}. 
However, the lack of parallel training data renders existing methods non-applicable to many text style transfer tasks. Instead of training with paired sentences, recent studies~\cite{fu2018style,hu2017toward,prabhumoye2018style,shen2017style,dianqiemnlp19} addressed this problem by using adversarial learning techniques. In this paper, we argue while the existing methods address the parallel data acquisition difficulty, they do not address the diversity problem in the translated outputs. We address the issue by formulating text style transfer as a one-to-many mapping problem and demonstrate one-to-many style transfer results.

\noindent{\bf Generative adversarial network (GANs)}~\cite{arjovsky2017wasserstein,goodfellow2014generative,salimans2016improved,CycleGAN2017} have achieved great success on image generation~\cite{huang2018multimodal,zhu2017toward}. Several attempts are made to applying GAN for the text generation task~\cite{guo2018long,lin2017adversarial,yu2017seqgan,zhang2017adversarial}. However, these methods are based on unconditional GANs and tend to generate context-free sentences. Our method is different in that our model is conditioned on the content and style codes, and our method allows a more controllable style transfer.

\section{Conclusion}
We have presented a novel framework for generating different style transfer outputs for an input sentence. This was achieved by modeling the style transfer as a one-to-many mapping problem with a novel latent decomposition scheme. Experimental results showed that the proposed method achieves better performance than the baselines in terms of the diversity and the overall quality.

\bibliography{nlp_style}
\bibliographystyle{acl_natbib}

\newpage
\appendix

\begin{table}[t]
\centering
\resizebox{\columnwidth}{!}{
    \begin{tabular}{lccccc}
    \toprule
	Method  & Style & Content & Fluency & Diversity & Overall\\
	\midrule
	Ours & 36.36 & 42.42 & \textbf{\textbf{51.52}} & \textbf{\textbf{72.22}} & \textbf{\textbf{51.01}}\\
	No Pref. & 28.79 & \textbf{\textbf{43.43}} & 35.35 & 16.16 & 36.87\\
    CAE$_{\sigma=0.001}$ & \textbf{\textbf{34.85}} & 14.14 & 13.13 & 11.62 & 12.12\\
	\midrule
	Ours & 30.30 & 42.93 & 41.41 & \textbf{\textbf{72.73}} & \textbf{\textbf{51.01}}\\
	No Pref. & \textbf{\textbf{35.35}} & \textbf{\textbf{43.94}} & \textbf{\textbf{44.95}} & 14.65 & 33.84\\
    CAE$_{\sigma=0.01}$ & 34.34 & 13.13 & 13.64 & 12.63 & 15.15\\
	\midrule
	Ours & 34.34 & \textbf{\textbf{43.94}} & \textbf{\textbf{48.48}} & \textbf{\textbf{60.10}} & \textbf{\textbf{47.47}}\\
	No Pref. & 29.29 & 42.93 & 42.42 & 28.79 & 39.39\\
    CAE$_{\sigma=0.1}$ & \textbf{\textbf{36.36}} & 13.13 & 9.09 & 11.11 & 13.13\\
	\midrule
	Ours & 24.24 & \textbf{\textbf{48.48}} & \textbf{\textbf{41.41}} & \textbf{\textbf{56.57}} & \textbf{\textbf{50.00}}\\
	No Pref. & 36.87 & 35.86 & 37.88 & 28.28 & 33.33\\
    CAE$_{\sigma=1}$ & \textbf{\textbf{38.89}} & 15.66 & 20.71 & 15.15 & 16.67\\
	\midrule
	Ours & 30.81 & \textbf{\textbf{44.95}} & 37.88 & \textbf{\textbf{48.99}} & \textbf{\textbf{44.44}}\\
	No Pref. & \textbf{\textbf{35.86}} & 41.41 & \textbf{\textbf{41.41}} & 39.39 & 34.34\\
    CAE$_{\sigma=10}$ & 33.33 & 13.64 & 20.71 & 11.62 & 21.21\\
	\midrule
	\midrule
	Ours & 33.84 & 42.42 & \textbf{\textbf{46.97}} & \textbf{\textbf{68.18}} & \textbf{\textbf{43.94}}\\
	No Pref. & 29.80 & \textbf{\textbf{48.80}} & 37.88 & 17.17 & 41.92\\
    CAE$_{k=\{1\}}$ & \textbf{\textbf{36.36}} & 9.09 & 15.15 & 14.65 & 14.14\\
	\midrule
	Ours & \textbf{\textbf{36.87}} &  43.43 & \textbf{\textbf{45.96}} & \textbf{\textbf{76.77}} & \textbf{\textbf{48.48}}\\
	No Pref. & 26.26 & \textbf{\textbf{45.96}} & 40.91 & 11.11 & 37.88\\
    CAE$_{k=\{1,5\}}$ & \textbf{\textbf{36.87}} & 10.61& 13.13 & 12.12 & 13.64\\
	\midrule
	Ours & 32.32 & 41.92 & \textbf{\textbf{44.44}} & \textbf{\textbf{71.72}} & \textbf{\textbf{46.46}}\\
	No Pref. & 26.77 & \textbf{\textbf{46.46}} & 41.41 & 15.15 & 38.38\\
    CAE$_{k=\{1,5,10\}}$ & \textbf{\textbf{40.91}} & 11.62 & 14.14 & 13.13 & 15.15\\
	\midrule
	Ours & 31.31 & \textbf{\textbf{43.94}} & \textbf{\textbf{50.51}} & \textbf{\textbf{73.74}} & \textbf{\textbf{47.47}}\\
	No Pref. & 33.33 & 43.43 & 36.87 & 12.63 & 13.64\\
    CAE$_{k=\{1,5, 10,15\}}$ & \textbf{\textbf{35.35}} & 12.63 & 12.63 & 13.64 & 15.66\\
	\bottomrule
\end{tabular}}
	\caption{Human preference comparison with the \textbf{CAE} on one-to-many style transfer results. The numbers are the user preference score of competing methods.}
	\label{tbl:user-cae}
\end{table}

\begin{table}[t]
\centering
\resizebox{\columnwidth}{!}{
\begin{tabular}{lccccc}
    \toprule
	Method  & Style & Content & Fluency & Diversity & Overall\\
	\midrule
Ours & 34.34 & 37.88 & \textbf{\textbf{43.94}} & \textbf{\textbf{66.67}} & \textbf{\textbf{43.43}}\\
	No Pref. & 30.30 & \textbf{\textbf{47.47}} & 42.93 & 22.22 & 40.40\\
    BTS$_{\sigma=0.001}$ & \textbf{\textbf{35.35}} & 14.65 & 13.13 & 11.11 & 16.16\\
	\midrule
	Ours & 37.88 & 38.38 & \textbf{\textbf{44.95}} & \textbf{\textbf{54.55}} & \textbf{\textbf{46.46}}\\
	No Pref. & 22.73 & \textbf{\textbf{45.45}} & 34.85 & 32.32 & 34.34\\
    BTS$_{\sigma=0.01}$ & \textbf{\textbf{39.39}} & 16.16 & 20.20 & 13.13 & 19.19\\
	\midrule
	Ours & 29.80 & \textbf{\textbf{42.42}} & \textbf{\textbf{45.96}} & \textbf{\textbf{50.51}} & \textbf{\textbf{50.51}}\\
	No Pref. & 29.29 & 41.92 & 36.36 & 35.86 & 35.35\\
    BTS$_{\sigma=0.1}$ & \textbf{\textbf{40.91}} & 15.66 & 17.68 & 13.64 & 14.14\\
	\midrule
	Ours & 33.33 & \textbf{\textbf{42.93}} & \textbf{\textbf{46.97}} & 38.89 & \textbf{\textbf{52.53}}\\
	No Pref. & 31.31 & 40.40 & 33.33 & \textbf{\textbf{46.97}} & 24.24\\
    BTS$_{\sigma=1}$ & \textbf{\textbf{35.35}} & 16.67 & 19.70 & 14.14 & 23.23\\
	\midrule
	Ours & 34.34 & \textbf{\textbf{50.51}} & \textbf{\textbf{55.56}} & \textbf{\textbf{63.64}} & \textbf{\textbf{59.60}}\\
	No Pref. & 30.30 & 33.84 & 25.25 & 25.25 & 18.69\\
    BTS$_{\sigma=10}$ & \textbf{\textbf{35.35}} & 15.66 & 19.19 & 11.11 & 21.72\\
	\midrule
	\midrule
	Ours & 31.31 & \textbf{\textbf{45.96}} & 41.41 & \textbf{\textbf{72.22}} & \textbf{\textbf{56.06}}\\
	No Pref. & 28.28 & 44.44 & \textbf{\textbf{42.93}} & 15.15 & 32.83\\
    BTS$_{k=\{1\}}$ & \textbf{\textbf{40.40}} & 9.6 & 15.66 & 12.63 & 11.11\\
	\midrule
	Ours & 37.88 & 39.39 & \textbf{\textbf{48.48}} & \textbf{\textbf{71.72}} & \textbf{\textbf{48.99}}\\
	No Pref. & 19.70 & \textbf{\textbf{48.99}} & 37.37 & 14.14 & 35.86\\
    BTS$_{k=\{1,5\}}$ & \textbf{\textbf{42.42}} & 11.62 & 14.14 & 14.14 & 15.15\\
	\midrule
	Ours & \textbf{\textbf{37.88}} & 36.87 & \textbf{\textbf{44.95}} & \textbf{\textbf{71.72}} & \textbf{\textbf{47.47}}\\
	No Pref. & 25.25 & \textbf{\textbf{47.47}} & 38.89 & 13.64 & 35.35\\
    BTS$_{k=\{1,5,10\}}$ & 36.87 & 15.66 & 16.16 & 14.65 & 17.17\\
	\midrule
	Ours & \textbf{\textbf{36.36}} & 44.95 & 41.92 & \textbf{\textbf{72.73}} & \textbf{\textbf{56.57}}\\
	No Pref. & 27.78 & \textbf{\textbf{46.46}} & \textbf{\textbf{45.96}} & 11.62 & 31.31\\
    BTS$_{k=\{1,5,10,15\}}$ & 35.86 & 8.59 & 12.12 & 15.66 & 12.12\\
	\bottomrule
\end{tabular}}
	\caption{Human preference comparison with the \textbf{BTS} on one-to-many style transfer results. The numbers are the user preference score of competing methods.}
	\label{tbl:user-bts}
\end{table}

\begin{table*}[h]
\centering
\resizebox{\textwidth}{!}{
\begin{tabular}{cccccc}
    \toprule
	Recon. Loss  & Back-Trans. Loss & Style Cls. Loss & Style Score & Content Preservation Score & BLEU\\
	\midrule
	\xmark  & \cmark & \xmark & $31.20$ & $45.89$ & $0.00 ~(66.2/0.2/0.0/0.0)$\\
	\xmark  & \xmark & \cmark & $75.30$ & $20.65$ & $0.00 ~(0.0/0.0/0.0/0.0)$\\
	\xmark  & \cmark & \cmark & $100.00$ & $42.46$ & $0.00 ~(35.8/0.1/0.1/0.0)$\\	
	\midrule
	\cmark  & \xmark & \xmark & $57.40$ & $90.10$ & $41.28 ~(69.7/48.2/34.8/25.2)$\\
	\cmark  & \cmark & \xmark & $61.38$ & $90.14$ & $39.72 ~(70.6/46.5/32.9/23.3)$\\
	\cmark  & \xmark & \cmark & $90.05$ & $74.84$ & $13.87~(48.0/20.4/9.6/4.1)$\\
	\cmark  & \cmark & \cmark & $82.64$ & $83.11$ & $24.5 ~(59.0/31.8/18.7/10.7)$\\
	\bottomrule
\end{tabular}
}
\caption{Empiricial analysis of the impact of each term in the proposed objective function for the proposed one-to-many style transfer task.}
\label{tbl:abalation}
\end{table*} 

{\large{\bf{Appendix}}}

\section{User Study}

To control the quality of human evaluation, we conduct pilot study to design and improve our evaluation questionnaire.
We invite 23 participants who are native or proficient English speakers to evaluate the sentences generated by different methods. 
For each participant, we randomly present $10$ sentences from Yelp500 test set, and the corresponding style transferred sentences generated by different models. We ask the participants to vote the transferred sentence which they think the sentence meaning is closely related to the original sentence with an opposite sentiment. 
However, we find that it may be difficult to interpret the evaluation results in terms of transfer quality in details.

\begin{figure*}[t]
	\centering
\includegraphics[width=0.99\textwidth]{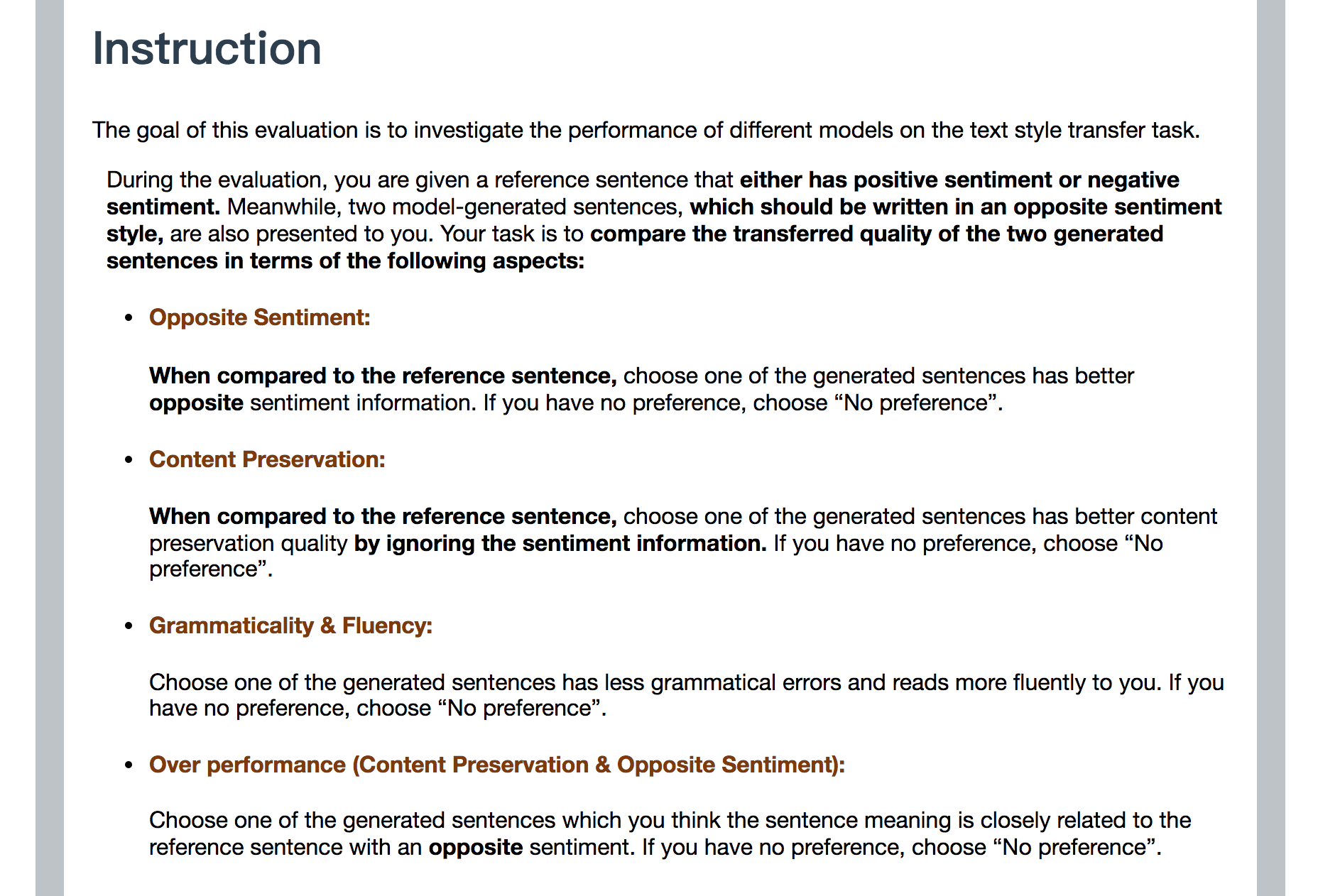}
	\caption{Instruction of our questionnaire on Amazon Mechanical Turk platform.}
	\label{fig:amt-intro}
\end{figure*}

\begin{figure*}[t]
	\centering
\includegraphics[width=.99\textwidth]{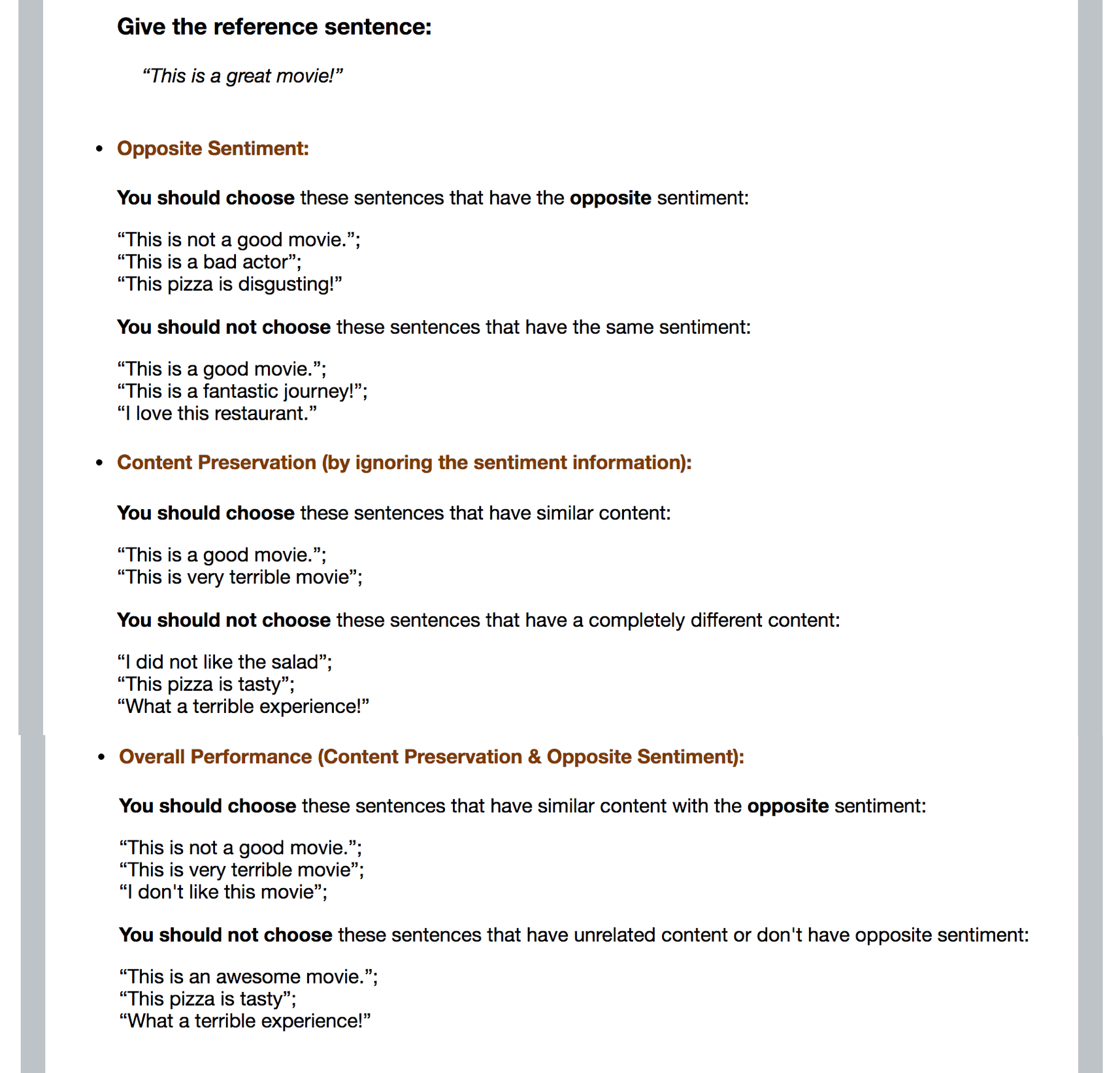}
	\caption{Example and guideline of our questionnaire on Amazon Mechanical Turk platform.}
	\label{fig:amt-exp}
\end{figure*}

Therefore, instead of asking the participants to directly
vote one sentence, we switch the task to evaluating the sentences in terms of four different aspects including style transfer, content preservation, fluency and grammatically, and overall performance. Following the literature~\cite{prabhumoye2018style}, for each pairwise comparison, a third option \textit{No Preference} is given for cases that both are equally good or bad. Figure~\ref{fig:amt-intro} and Figure~\ref{fig:amt-exp} show the instructions and the guidelines of our questionnaire for human evaluation on Amazon Mechanical Turk platform. We refer the reader to Section 3 in the main paper for the details of the human evaluation results.

To evaluate the performance of one-to-many style transfer, we extend the pair-wise comparison to set-wise comparison. Given an input sentence and two sets of model-generated sentences (5 sentences per set), the workers are asked to choose which set has more diverse sentences with the same meaning, and which set provides more desirable sentences considering both content preservation and style transfer. We also ask the workers to compare the transfer quality in terms of content preservation, style transfer, grammatically and fluency.

\section{Diversity Baselines}

We report further comparisons with different variants of \textbf{CAE} and \textbf{BTS}. We added random Gaussian noise to the style code of \textbf{CAE} and \textbf{BTS}, respectively. Specifically, we randomly sample the noise from the Gaussian distribution with $\mu=0$ and $\sigma\in\{0.001, 0.01, 0.1, 1, 10\}$, respectively. We empirically found that the generations will be of poor quality when $\sigma>10$. Thus, we evaluated the baselines with $\sigma\leq10$ in the experiments. On the other hand, we also explored different extensions to enhance the diversity of sequence generation of the baselines. For example, we expanded the generations by randomly select a beam search size $k\in\{1,5,10,15\}$ per generation.

\section{Additional One-to-Many Style Transfer User Study Results}

We report the human evaluation with comparisons to different variants of the \textbf{CAE} and \textbf{BTS}. Similar to the human study presented in the main paper, we conduct evaluation using Amazon Mechanical Turk. We randomly sampled $200$ sentences from Yelp test set for user study. Each comparison is evaluated by at least three experts whose HIT Approval Rate is greater than $90\%$. We received more than $3600$ responses, and the results are summarized in Table~\ref{tbl:user-cae} and Table~\ref{tbl:user-bts}. We observed previous models achieve higher style scores, but their output sentences are often in a generic format and may not preserve the content with correct grammar. In contrast, our method achieves significantly better performance than the baselines in terms of diversity, fluency, and overall quality. 

\section{Ablation Study on Objective Function}

The proposed objective function consists of five different learning objectives. We conduct ablation study to understand which loss function contributes to the performance. Since adversarial loss is essential for domain alignment, we evaluate loss functions by iterating different combination of the reconstruction loss, the back-translation loss (together with the mean square loss), and the style loss. 

We report the style score and the content preservation score in this experiment. We additionally present the BLEU score~\cite{papineni2002bleu}, which is a common metric for evaluating the performance of machine translation. A model with a higher BLEU score means that the model is better in translating reasonable sentences. As shown in Table~\ref{tbl:abalation}, we find that training without reconstruction loss may not produce reasonable sentences according to the BLEU score. Training with reconstruction loss works well for content preservation yet it performs less favorably for style transfer. Back-translation loss is able to improve style and content preservation scores since it encourage content and style representations to be disentangle. When training with the style loss, our model improves the style accuracy, yet performs worse on content preservation. Overall, we observe that training with all the objective terms achieves a balanced performance in terms of different evaluation scores. The results show that the reconstruction loss, the back-translation loss, and the style loss are important for style transfer. 

\section{Style Code Sampling Scheme}
We design a sampling scheme that can lead to a more accurate style transfer. During inference, our network takes the input sentence as a query, and retrieves a pool of target style sentences whose content information is similar to the query. We measure the similarity by estimating the cosine similarity between the sentence embeddings. Next, we randomly sample a target style code from the retrieved pool, and generate the output sentence. 
The test-time sampling scheme improves the content preservation score from $83.11$ to $83.41$, and achieves similar style score from $82.64$ to $82.66$ on Yelp25000 test set. The results show that it is possible to improve the content preservation by using the top ranked target style sentences.

\begin{figure}[t]
	\centering
\includegraphics[width=.99\columnwidth]{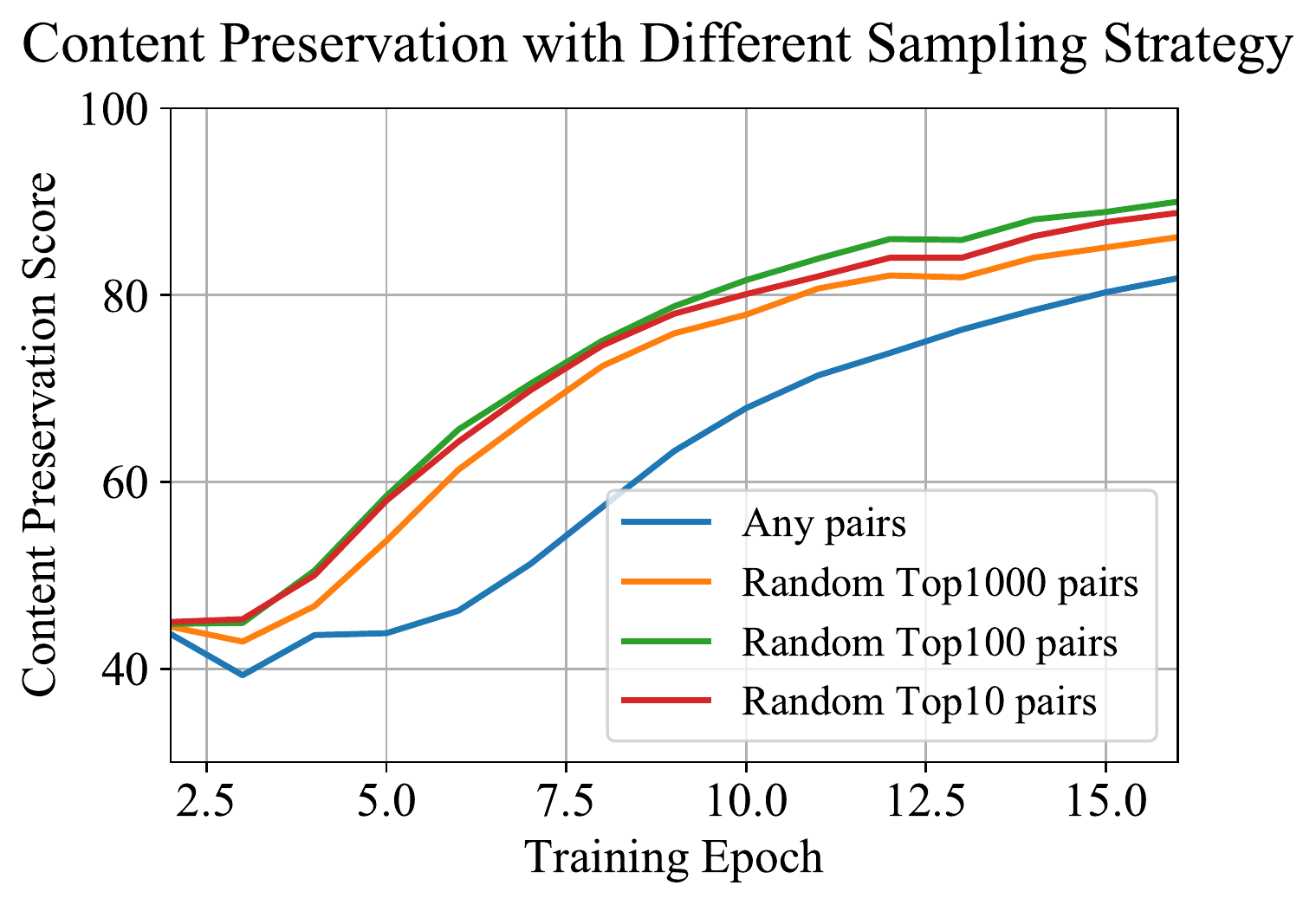}
	\caption{Performance comparison of our model using different sampling schemes.}
	\label{fig:content_nn}
\end{figure}

\begin{figure}[t]
	\centering
\includegraphics[width=.95\columnwidth]{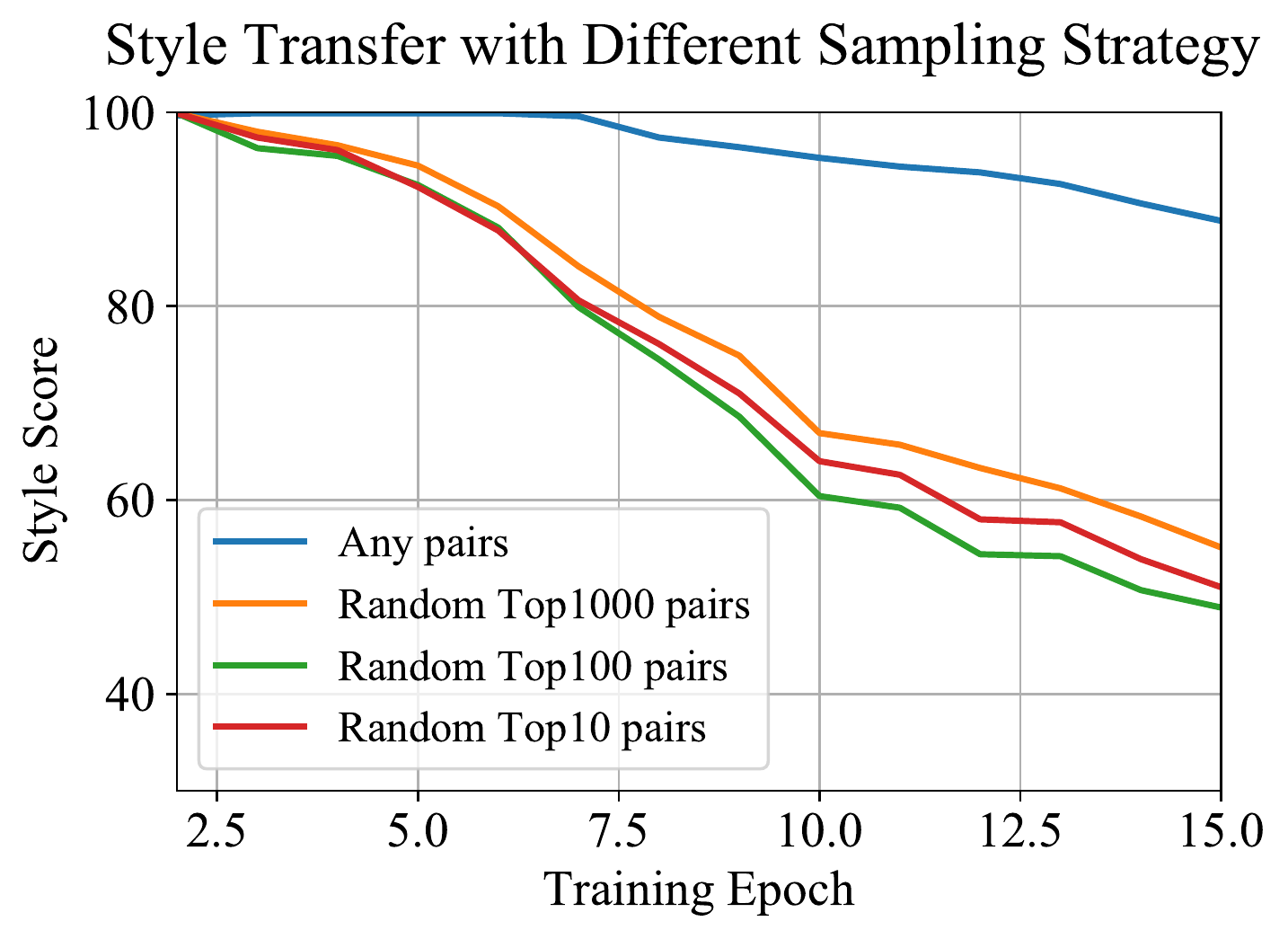}
	\caption{Performance comparison of our model using different sampling schemes.}
	\label{fig:style_nn}
\end{figure}

We provide further analysis on the sampling scheme for the training phase. Specifically, during training, we sample the target style code from the pool of top ranked sentences in the target style domain. Figure~\ref{fig:content_nn} shows the content preservation scores of our method using different sampling schemes. The results suggest we can improve the content preservation by learning with the style codes extracted from the top ranked sentences in the target style domain. However, we noticed that this sampling scheme actually reduces the number of training data. It becomes more challenging for the model to learn the style transfer function as shown in Figure~\ref{fig:style_nn}. The results suggest that it is more suitable to apply the sampling scheme in the inference phase.

\section{Additional Implementation Details}
We use $256$ hidden units for the content encoder, the style encoder, and the decoder. All embeddings in our model have dimensionality $256$. We use the same dimensionalities for linear layers mapping between the hidden and embedding sizes. Additionally, we modify the convolution block in the style encoder $E_i^s$ to have max pooling layers for capturing the activation of the style words. On the other hand, we also modify the convolution block of the content encoder $E_i^c$ to have average pooling layers for computing the average activation of the input. During inference, the decoder generates the output sentence with the multi-step attention mechanism~\cite{gehring2017convolutional}. 

\begin{table}
\centering
\fbox{
\begin{minipage}{18em}
\raggedright
\textbf{Input:} I stayed here but was disappointed as its air conditioner does not work properly.\\
\textbf{Output:} I love here but was but as well feel's me work too.\\
\rule[0.3\baselineskip]{\textwidth}{0.05pt}
\textbf{Input:} I might as well been at a street fest it was so crowded everywhere.\\
\textbf{Output:} I well as well a at a reasonable price it was so pleasant.\\
\rule[0.3\baselineskip]{\textwidth}{0.05pt}
\textbf{Input:} Free cheese puff - but had rye in it (I hate rye!).\\
\textbf{Output:} It's not gourmet but it definitely satisfies my taste for good Mexican food.\\
\end{minipage}
}
\caption{Example failure cases generated by the proposed method.}
\label{tbl:error}
\end{table}


\section{Failure Cases}
Although our approach performs more favorably against the previous methods, our model still fails in a couple of situations. Table~\ref{tbl:error} shows the common failure example generated by our model. We observe that it is challenging to preserve the content when the inputs are the lengthy sentences. It is also challenging to transfer the style if the sentence contains novel symbols or complicated structure.

\end{document}